# DW-A-PRM: A Dynamic Weighted Planner

Siyuan Wang[*,#]
Nanjing University of Science and Technology ZiJin College
Nanjing, China
19852225218@163.com

Shuyi Zhang[#]
Nanjing University of Science and Technology ZiJin College
Nanjing, China

Zhen Tian[#]
University of Glasgow
Glasgow, UK

Yuheng Yao[#]
Nanjing University of Science and Technology ZiJin College
Nanjing, China

GongSen Wang
Nanjing University of Science and Technology
Nanjing, China

Yu Zhao
Nanjing University of Science and Technology ZiJin College
Nanjing, China

[#]These authors contributed equally.

***Abstract*—Robot path planning plays a pivotal role in enabling autonomous systems to navigate safely and efficiently in complex and uncertain environments. Despite extensive research on classical graph-based methods and sampling-based planners, achieving an optimal balance between global optimality, computational efficiency, and adaptability to dynamic environments remains an open challenge. To address this issue, this paper proposes a hybrid path planning framework, which integrates heuristic-driven search with probabilistic roadmap construction under a dynamic weighting scheme. By coupling the global guidance of A* with the stochastic exploration of PRM, the method achieves a synergistic balance between search optimality and computational tractability. Comprehensive experiments in diverse simulated environments demonstrate that the proposed method consistently yields smoother and shorter paths while significantly reducing computational overhead compared with conventional approach and other hybrid planners. These results highlight the potential of the proposed framework as an effective and generalizable solution for real-time robotic navigation in complex environments.**

## I. INTRODUCTION AND RELATED WORKS

### *A. Introduction: Path Planning for indoor environment*

In recent years, intelligent robots have become deeply integrated into numerous aspects of daily life and industrial production. As one of the most essential capabilities of such systems, autonomous navigation has gained increasing prominence in applications that span from self-driving vehicles and aerial drones to field exploration and emergency response. The continual evolution of sensing and perception technologies has further accelerated this trend—modern robots now leverage advanced systems such as cameras, inertial sensors, and real-time mapping modules to perceive and interpret their surroundings with remarkable accuracy and adaptability. Path planning aims to generate collision-free trajectories for avoiding obstacles [1]. It has been extensively applied in intelligent transportation [2], bio-robotics [3], and industrial automation [4]. Efficient algorithms play a crucial role in multiple domains [5].

In a mobile robotic system, the overall navigation process is typically structured around three interconnected modules: perception, decision-making, and planning. The perception layer serves as the robot’s sensory interface with the environment, continuously gathering and interpreting data from cameras, LiDAR, or other onboard sensors to build a coherent understanding of its surroundings [6]. The decision-making layer then reasons over this representation to assess navigational risks, predict environmental changes, and select feasible motion intentions. Finally, the planning layer synthesizes these intentions into a smooth and collision-free trajectory that respects the robot’s kinematic and dynamic constraints.

This hierarchical flow highlights the continuous interaction between sensing, cognition, and motion generation. Beginning with raw sensory perception, the robot identifies structural and dynamic features of the environment; through decision-making, it transforms this understanding into actionable movement choices; and through planning, it converts these choices into an executable path toward the goal. Such integration enables robots to operate reliably in complex, cluttered, and dynamically changing environments. Since the introduction of Simultaneous Localization and Mapping (SLAM) technology [7,8], robots have been able to simultaneously construct environmental maps and estimate their own positions in unknown spaces, laying the foundation for modern autonomous navigation. With the evolution of this technology, global path planning emerged as a central component of navigation systems in the 21st century. As a single-query optimization problem, it seeks an efficient and collision-free route from the start to the goal while addressing spatial complexity and temporal uncertainty in dynamic environments. features. SLAM technology enabled robots to effectively localize and build maps in previously unknown environments, significantly advancing navigation technology [9]. By the 21st century, navigation algorithms diversified, with the design of global path planning algorithms becoming pivotal to the execution navigation tasks [10]. Typically regarded as a single-query planning problem, global path planning aims to feasible path from the robot's workspace tothetarget,ensuring that the local planning algorithm executes rapidly[11]. This task involves searching for the optimal pathfromthe starting point to the destination in environments that maybe unknown or dynamic, addressing challenges relatedtospatial complexity and temporal uncertainty [12].

Extensive research has advanced the field of path planning [13]. Classical graph-based methods, such as Dijkstra’ s algorithm [14], guarantee globally optimal paths but suffer from high computational cost, while A [15] improves efficiency through heuristics yet remains sensitive to heuristic design and environmental changes. Sampling-based approaches, including PRM and RRT, enhance scalability in high-dimensional spaces but face challenges in dynamic updates, path smoothness, and the curse of dimensionality. Improved variants like RRT and JPS further optimize path quality and grid efficiency but struggle with real-time adaptability. Hybrid planners, such as Lazy-PRM and RRT-A*, attempt to integrate deterministic search with stochastic exploration; however, their fixed heuristics and static sampling limit robustness in complex, dynamic environments.

This work introduces a hybrid path planning framework, termed A-PRM*, that integrates heuristic search with probabilistic sampling through a dynamic fusion strategy. The approach aims to overcome two key limitations of conventional planners using sampling bias in PRM leading to poor path feasibility, and the high computational cost of A* in complex, high-dimensional spaces.

The proposed framework operates through a three-stage cooperative process. In the heuristic roadmap construction phase, the A* heuristic is embedded into the PRM sampling process to adaptively regulate node density according to obstacle distribution, producing a roadmap with nonuniform spatial granularity. In the adaptive graph search phase, this roadmap serves as the search domain for a dynamically weighted A* algorithm that balances exploration efficiency and path optimality in real time. Finally, a path refinement phase applies smoothing and local adjustment to reduce discontinuities introduced by random sampling.

Extensive experiments across diverse environments demonstrate that A*-PRM substantially improves planning reliability, generates shorter and smoother paths, and significantly reduces computational burden compared with classical PRM and hybrid baselines. These results highlight the algorithm’s robustness and suitability for real-world autonomous navigation tasks.

### *B. Related Works: Path Planning for Indoor Environment*

Path planning in indoor environments is challenging due to confined spaces, dense obstacle layouts, and frequent environmental changes. Classical algorithms such as Dijkstra and A* provide reliable shortest-path solutions but often face computational inefficiency in large or dynamic indoor maps. The Artificial Potential Field (APF) method introduces a physics-inspired mechanism that models obstacles and goals as repulsive and attractive forces, respectively; however, it is prone to local minima in complex spaces.

To overcome these limitations, sampling-based algorithms such as PRM and RRT have been widely adopted. These methods explore the configuration space through random sampling, offering better adaptability to high-dimensional environments. Recent research further extends these techniques using heuristic-guided sampling, adaptive connection strategies, and learning-assisted local planners, improving performance in cluttered indoor settings. Additionally, optimization-based frameworks such as Model Predictive Control (MPC) [16,17] and trajectory optimization approaches refine path smoothness and kinematic feasibility, which are essential for service and warehouse robots operating in constrained indoor environments.

### *C. Related Works: Path Planning for Public Transportation*

Path planning for public transportation focuses on large-scale coordination, multi-agent scheduling, and dynamic route optimization [18,19].

Modern research trends integrate optimization-based planning [20] and sampling-based [21] and multi-objective optimization. Techniques such as Mixed-Integer Linear Programming (MILP), Nonlinear MPC (NMPC), and multi-agent reinforcement learning (MARL) are increasingly applied to optimize travel time, passenger distribution, and congestion avoidance. By leveraging real-time data from GPS, IoT [22] sensors, and traffic forecasts, these hybrid frameworks enable adaptive and data-driven decision-making, forming the foundation for intelligent transportation networks and autonomous fleet management in urban transit systems.

## II. Related Approaches

### *A. A* Approach*

The A* is a heuristic graph search method that estimates the optimal path by combining actual travel cost with an admissible heuristic function. It effectively balances exploration and efficiency, ensuring near-optimal solutions in structured environments.

### *B. PRM Approach*

The PRM employs random sampling to construct a connectivity graph in the configuration space. By linking feasible nodes, it provides a scalable approach to global path planning, particularly in high-dimensional or complex environments.

### *C. A*-PRM Approach*

The proposed A*-PRM framework integrates the heuristic guidance of A* with the stochastic exploration of PRM. This hybrid design unites global search efficiency with sampling adaptability, forming a dynamic and flexible planning mechanism capable of handling complex obstacle distributions and diverse environmental conditions.

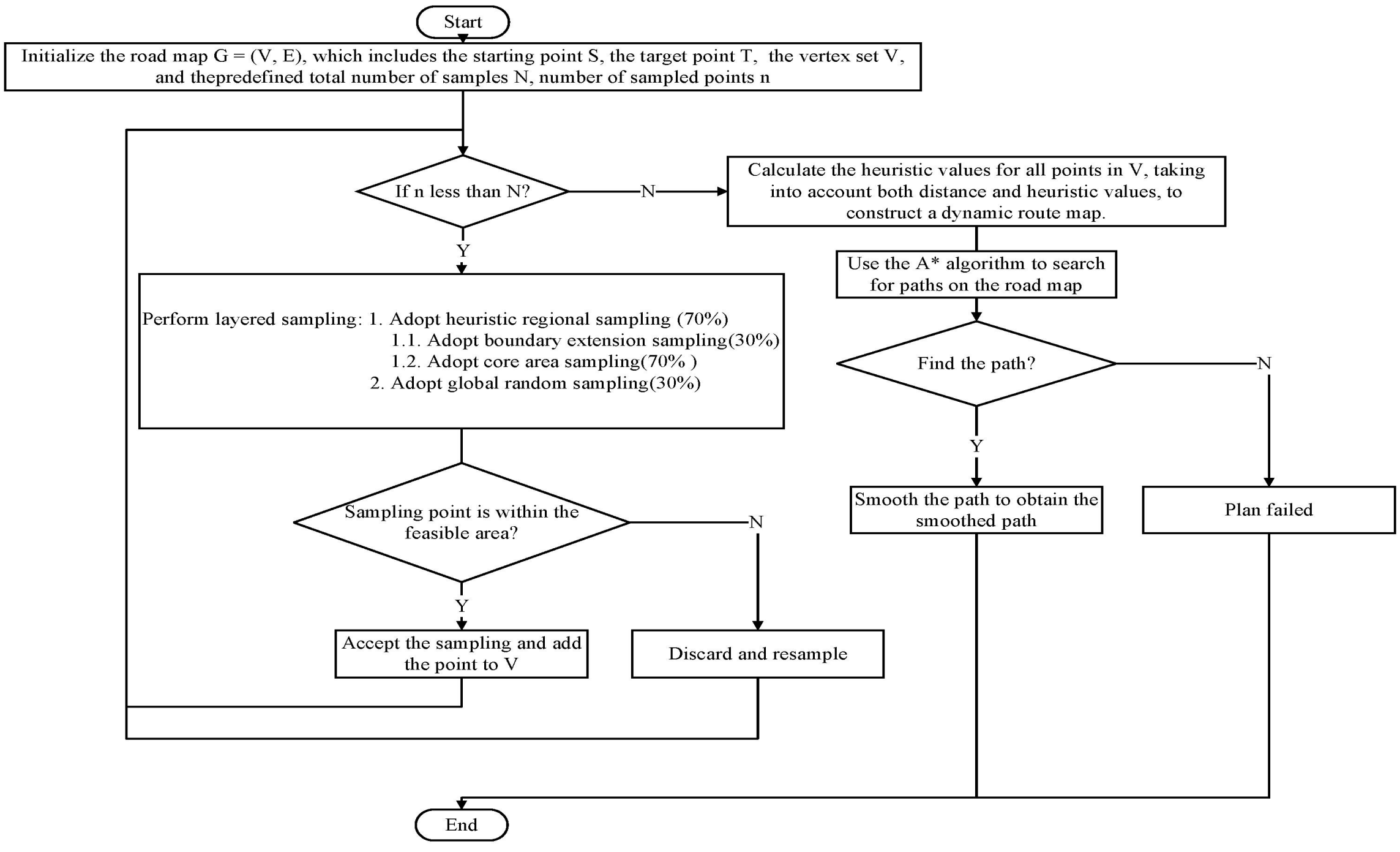

Fig. 1: Flowchart of the A*-PRM algorithm

(1) Heuristic Probabilistic Roadmap Construction

In this stage, nodes are generated using a guided sampling method.:

1) Node Initialization:

The start point S and goal point G are first added to the node set V. Then, most of the other nodes (about 7/10) are placed along the main path area between S and G, while the remaining are spread across the whole free space. This way, the roadmap focuses on the main route but still explores the entire environment.

2) Collision Monitoring:

Each node is checked against the map M to ensure it is in a collision-free area. Only safe nodes $C_{free}$ are kept..

3) Dynamic topology construction:

During graph construction, a heuristic-based dynamic connection mechanism is used to optimize the roadmap intelligently. For every single node $u$ , its $T$ heuristic distance to the goal is first computed by $h_u = heuristic(u,T)$, and $h_{\max} = heuristic(S,T)$. Then, the number of neighboring connections is adjusted dynamically according to the global maximum heuristic value, allowing nodes closer to the goal to form denser connections $k = \max\left[k_{neighbors} \times \left(1 - \frac{h_u}{h_{\max}}\right), 3\right]$. This mechanism reduces redundant connections for nodes near the starting point (where the heuristic value is larger) while increasing the connection density for nodes closer to the goal (where the heuristic value approaches zero). As a result, a highly directional network structure is formed. The selection of candidate neighbors follows a multi-objective optimization strategy that jointly considers both the Euclidean distance between nodes and the alignment of their heuristic directions. The proposed scoring function is computed as following equation:

$$\text{score} = \text{dist}(u,v) + 0.5 \times \left| heuristic(v,T) - h_u \right| \quad (1)$$

Nodes that are spatially nearing and hold heuristic values similar to the goal are tightly connected. Edges that with no collision risks are finally used, with intelligent weights using Euclidean distance which is a rational and usually item.

(2) Heuristic path definition

In this phase, the improved A* algorithm is executed on the constructed topological graph:

1)initialization

The actual cost of the start node is set to zero, and its total evaluation value is initialized accordingly $f_{score}[S] = heuristic(S,T)$ , $g_{score}$ . All other nodes $f_{score}$ are initialized with infinite cost values.

2)Priority Queue Extension

A priority queue is maintained in ascending order of node evaluation values, and the following operations are repeated:

a) Extract the node $u$ with the smallest evaluation value $f_{min}$ from the queue $Q$ .

b) If this node is the goal ( $u = G$ ), backtrack through its parent nodes to reconstruct the path $P$ .

c) Otherwise, traverse all neighboring nodes $u$ of $v$ and compute the temporary cost $g_{temp} = g(u) + \| u - v \|$ .

d) If the temporary cost is smaller than the recorded cost, update the parent node, the cost values, and the priority queue accordingly.

(3) Path post-processing

This phase refines the generated path to enhance its smoothness and feasibility.

1) Redundant Node Removal: Consecutive nodes that can be connected by a collision-free straight line are merged.

2) Smoothing: The optimized path is further smoothed using continuous functions to satisfy kinematic constraints and ensure easier control tracking.

## III. MODEL AND EXPERIMENT

### A. Description of the robot's complex scene route planning problem

A. Problem Description

In complex environments with dense or dynamic obstacles, robot path planning aims to achieve safe and efficient navigation under multiple constraints. The problem involves high-dimensional configuration spaces (position, velocity, and orientation) and competing objectives, such as maintaining safety margins (more than 0.5 m from obstacles) while ensuring smooth motion (maximum curvature less than 0.3 $m^{-1}$).

This task can be formulated as a mixed-integer nonlinear programming (MINLP) problem to balance safety, efficiency, and smoothness in complex scenes.

$$
\begin{cases}
C_{free} = \{ q \in \mathbb{R}^n \mid \Phi(q) = \sum_{k=1}^{K} \exp(-\frac{\|q - o_k\|^2}{2\sigma^2}) < \tau \} \\
\min_{\tau \in \mathrm{T}} \left[ w_1 \underbrace{\int_0^T \|\tau'(t)\| dt}_{path_length} + w_2 \underbrace{\max_t \frac{1}{d(\tau(t), \partial C_{obs})}}_{safety_margin} + w_3 \underbrace{\int_0^T \kappa^2(t) dt}_{smoothness} \right] \\
\kappa(t) = \frac{\|\tau'(t) \times \tau''(t)\|}{\|\tau'(t)\|^3} \\
\Omega_{dyn} = \{ \tau(t) \mid \forall t \in [0, T], \Phi(\tau(t), t) < \tau_{dynamic} \} \\
E(q_i, q_j) = \begin{cases} 1 & \text{if } \frac{\|q_j - q_{goal}\|}{\|q_i - q_{goal}\|} < \gamma \text{ and } \int_0^1 \Phi(q_i + s(q_j - q_i)) ds < \epsilon \\ 0 & \text{otherwise} \end{cases} \\
f(q) = g(q) + \eta h(q) + \mu \sum_{k=1}^{m} \lambda_k \phi_k(q)
\end{cases}
\quad (2)
$$

## IV. EXPERIMENT AND COMPARISON

### A. Experimental environment configuration

#### A1. Experimental Environment

All experiments were conducted on a Python 3.12.4 platform with an Intel Core i9-14900HX processor and 32 GB RAM. The test map is a 2D static warehouse grid (897 × 550 pixels), with the start at (50, 50) and the goal at (847, 500).

Both PRM and A*-PRM algorithms share the same base settings for fair comparison. The default number of sampled nodes is 1000, with additional tests using 500 and 3000 samples to analyze scalability. The neighbor connection number k is set to 10 (fixed for PRM, dynamically adjusted for A*-PRM), with comparative tests at k = 5 and k = 20.

Each experiment was repeated 12 times with identical maps and random seeds, and the average result was computed after removing the maximum and minimum values. The obstacle distribution and layout of the test map are illustrated in Fig. 2.

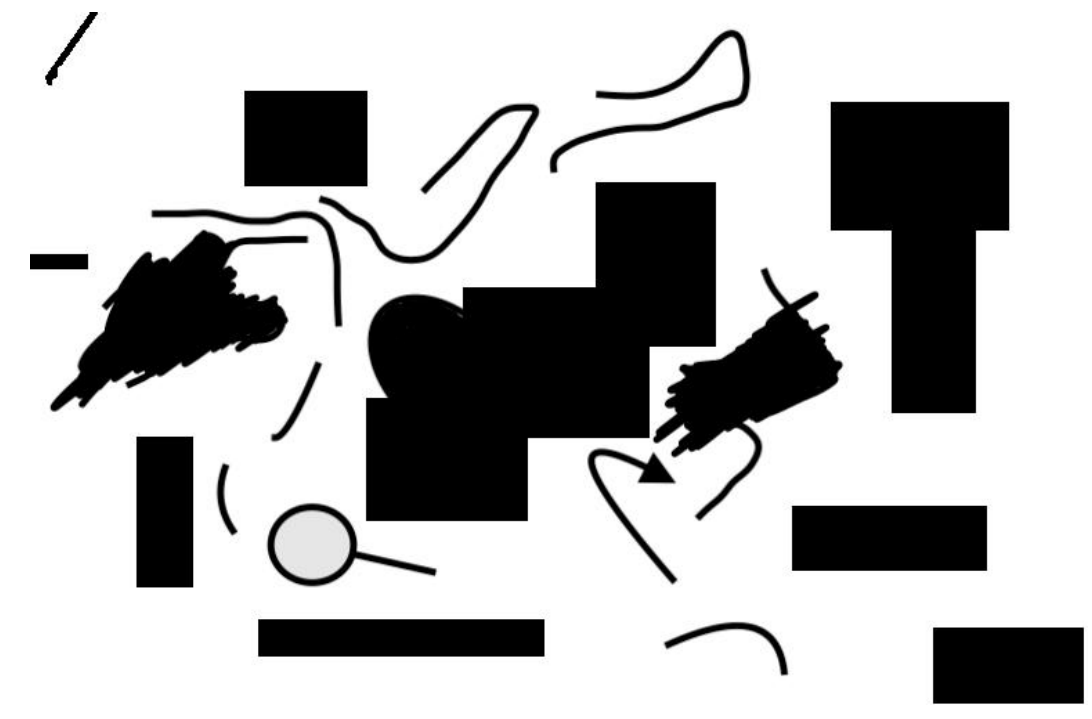

Fig.2: Grid map structure

### B. Experimental Results Analysis

#### B1. Evaluation criteria

This experiment uses four-dimensional quantitative indicators to evaluate algorithm performance, focusing on path optimality, computational efficiency and resource consumption.

a. Path length: Euclidean distance integral of continuous paths (unit: meter);

b. Running time: the total time taken from algorithm initialization to outputting a feasible path (in seconds);

c. Smoothness: average angle change (unit: degrees per meter);

d. Number of explored nodes: the total number of nodes visited by the algorithm during the search process (including sampling points and search nodes).

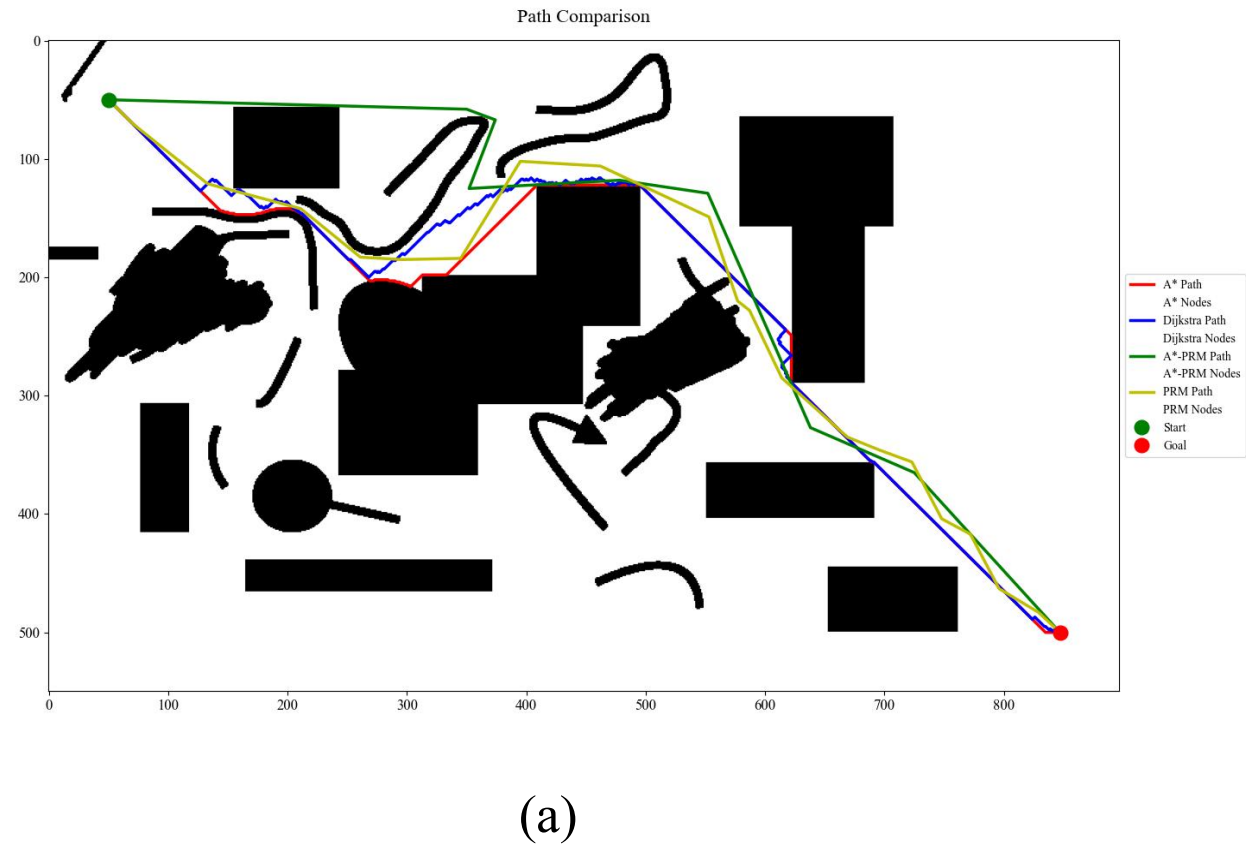

(a)

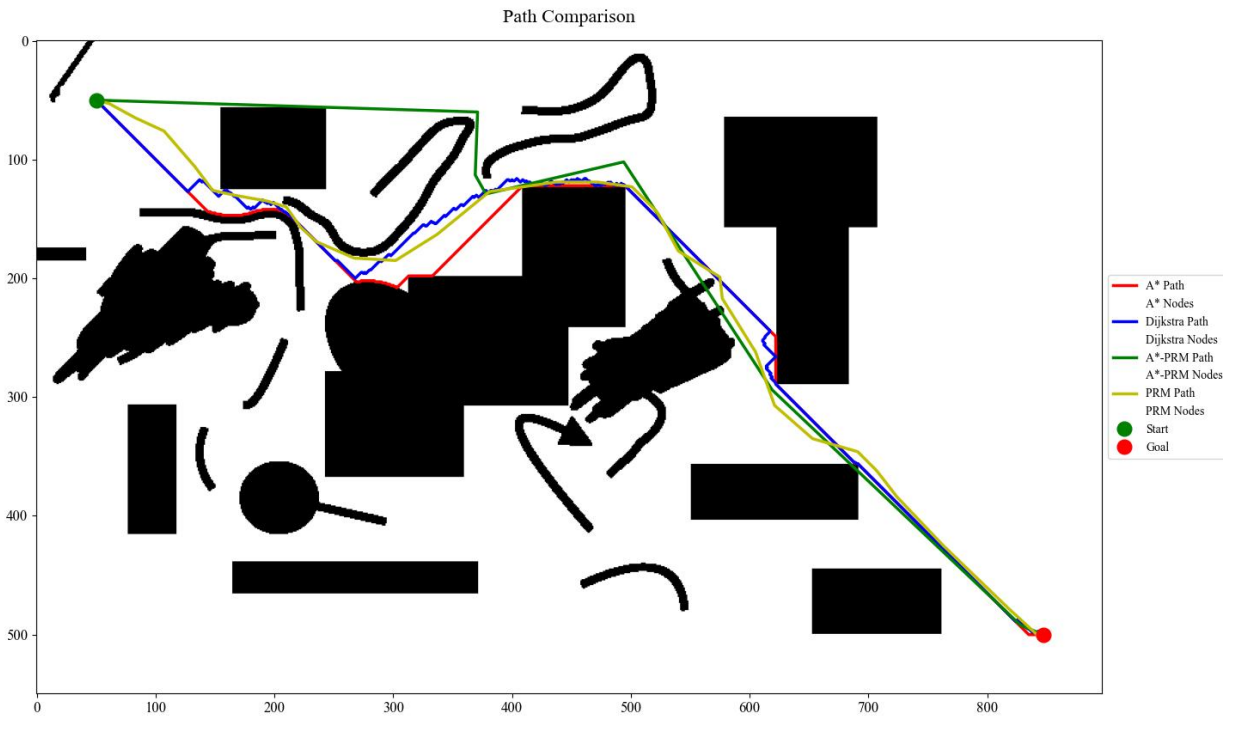

(b)

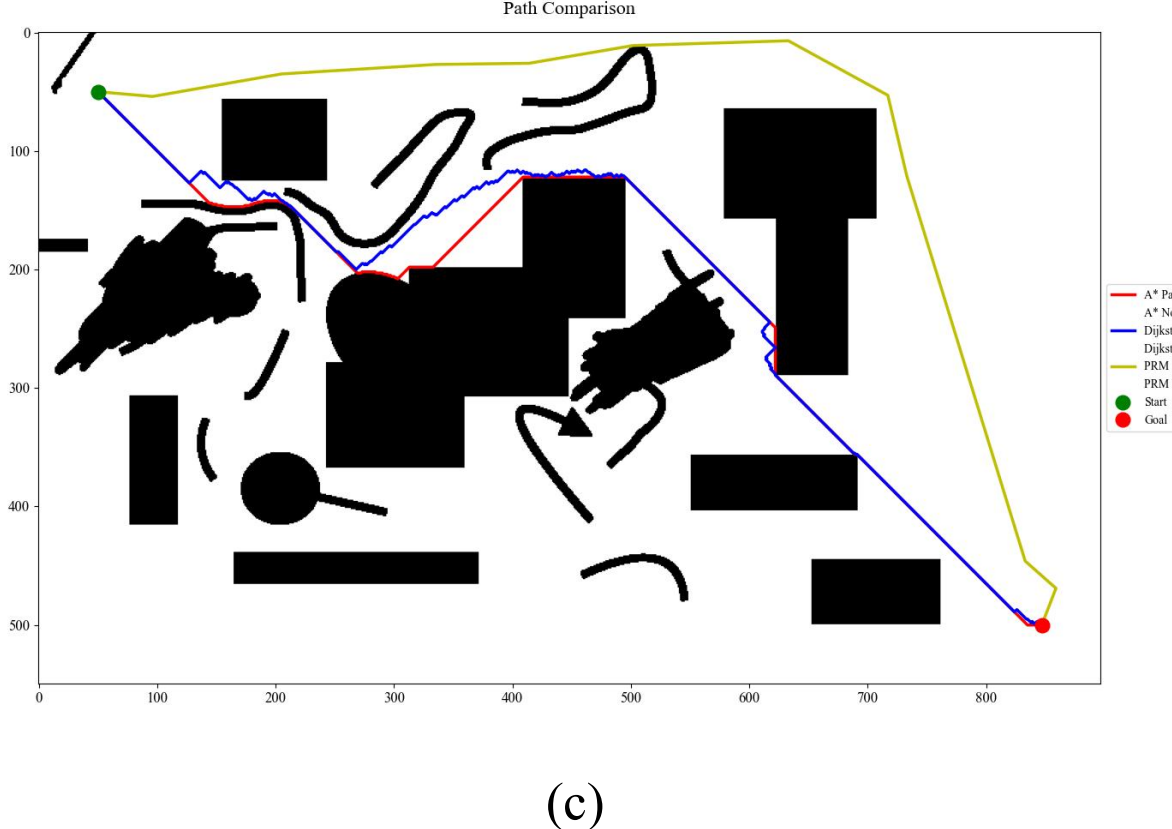

(c)

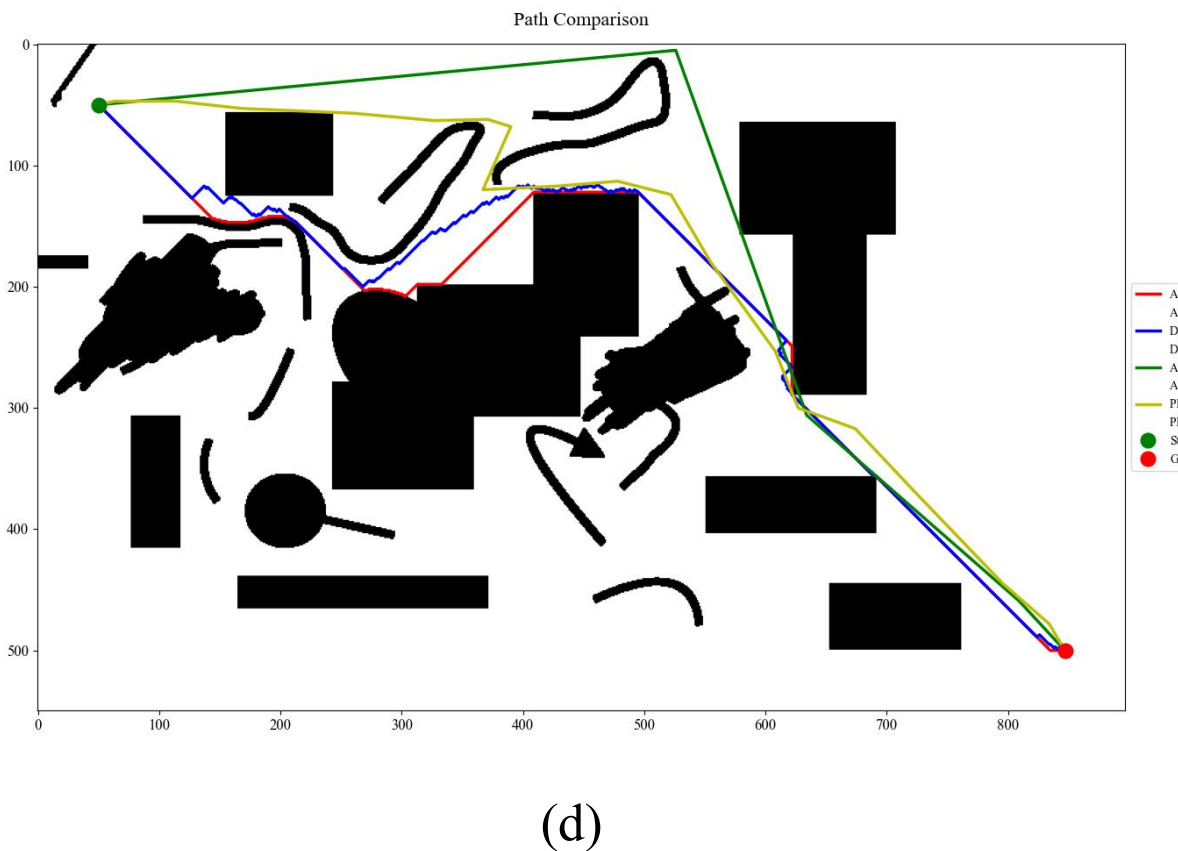

(d)

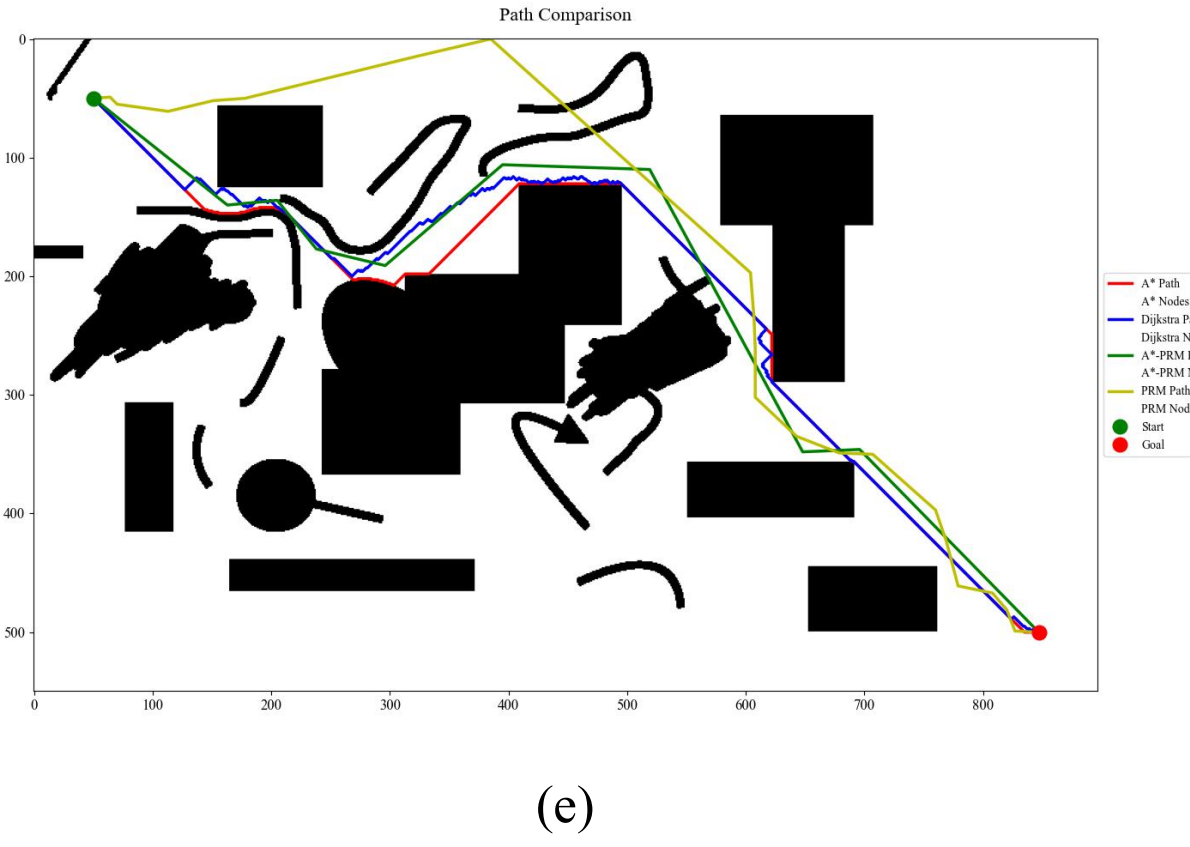

(e)

Fig.3: Path Planning results under various conditions

As shown in Fig. 3, the performance of the A*-PRM algorithm was evaluated under different combinations of sampling nodes (|V| = 500, 1000, 3000) and neighbor connections (k = 5, 10, 20). The results indicate that higher sampling densities yield smoother and more stable paths, while lower densities still maintain a high success rate due to the heuristic guidance mechanism. However, when k falls below a critical level, path continuity deteriorates, suggesting that an appropriate balance between sampling density and connectivity is essential. These findings highlight the robustness and adaptability of the A*-PRM algorithm across varying parameter settings.

## V. CONCLUSION AND OUTLOOK

This study proposes A-PRM*, a probabilistic roadmap algorithm enhanced with dynamic weighting. By integrating heuristic roadmap construction, adaptive graph search, and path post-optimization into a unified three-stage framework, the method combines the global exploration strength of PRM with the directed search efficiency of A*. Experimental evaluations show that the proposed algorithm achieves higher path stability, stronger adaptability to complex obstacle distributions, and better computational efficiency compared to traditional PRM.

The results suggest that A*-PRM is particularly suitable for high-precision navigation tasks in structured industrial environments, such as warehouse automation and robotic machining, where both path smoothness and stability are critical. Despite its advantages, the algorithm still faces challenges in adapting to dynamic environments, ensuring path smoothness within a closed optimization loop, and improving computational scalability. Future work will focus on developing adaptive sampling mechanisms for dynamic obstacle handling, introducing smoothness-aware optimization in the connection phase, and implementing GPU-accelerated frameworks to enhance real-time performance.